\documentclass{esannV2}
\usepackage{graphicx}
\usepackage{amssymb,amsmath,array}
\usepackage{multirow}
\usepackage{booktabs}
\usepackage{makecell}
\usepackage{url}
\usepackage{hyperref}
\usepackage{subcaption}

\begin{document}
\title{One Adapter, One Pass: Are Modular Adapters Necessary for Class-Incremental Learning?}

\author{Alejandro Dopico-Castro, Oscar Fontenla-Romero, Bertha Guijarro-Berdi\~nas, \\ Amparo Alonso-Betanzos
\thanks{
Under review at ESANN 2027 (European Symposium on Artificial Neural Networks, Computational Intelligence and Machine Learning). Funded by MICIU/AEI/10.13039/501100011033, ERDF/EU and ESF+/EU (PID2023-147404OB-I00), Horizon Europe (101070381), the Ministry for Digital Transformation and Civil Service and NextGenerationEU/PRTR (TSI-100925-2023-1) and Xunta de Galicia (ED431C 2026/38). CITIC, in the CIGUS Network, is funded by Xunta de Galicia and ERDF Galicia 2021-2027 (ED431G 2023/01).
}
\vspace{.3cm}\\
Universidade da Coru\~na, CITIC, Facultade de Inform\'atica \\ Campus de Elvi\~na s/n, A Coru\~na, Spain
}

\maketitle

\begin{abstract}
Pre-trained models ease class-incremental learning, in which new classes must be learned over time without forgetting previous ones, since a frozen backbone can be complemented with a lightweight adapter and an analytic classifier. However, if the adapter keeps learning, the statistics stored by the classifier become obsolete, and a growing trend instead isolates knowledge in additional adapters. In this work, we question whether multiple adapters are necessary. We show that a substantial part of the forgetting of a single adapter that keeps learning stems from classifier obsolescence rather than from a loss of representation quality. Once this obsolescence is corrected through closed-form drift compensation and feature distillation, a single adapter outperforms multi-adapter methods on all three benchmarks, with one forward pass instead of one per task and about $11\times$ less energy consumption. Finally, a controlled study shows that isolating knowledge in separate adapters only helps when tasks are clearly distinct; otherwise, even ideal isolation suffers from classifier obsolescence, whereas a single adapter remains stable.
\end{abstract}

\section{Introduction}

In real applications, data are rarely static and new categories may emerge long after a model has been deployed. Since retraining from scratch is expensive and past data may be unavailable, class-incremental learning (CIL) aims to incorporate new classes sequentially without forgetting previously learned ones. The classes arrive in a sequence of \emph{tasks}, each introducing a set of new classes, and the model must then classify all the classes seen so far. Large pre-trained models have made this goal more attainable~\cite{zhou2024aper}, as their representations generalise across domains and require only light adaptation.

A common recipe pairs a frozen backbone and a lightweight adapter with an analytic classifier~\cite{fontenla2021rolann}. Such a classifier updates accumulated feature statistics in closed form, which avoids the forgetting of gradient-based updates, but only for fixed features. If the adapter keeps learning, the stored statistics describe features that the model no longer produces, and the classifier becomes obsolete. The adapter is therefore usually frozen after the first task~\cite{zhou2024aper}, which avoids this obsolescence but prevents later tasks from improving it.

An alternative is to isolate knowledge in additional adapters, adding one adapter per task, as in EASE~\cite{zhou2024ease} and TUNA~\cite{wang2025integrating}, or expanding the model when a distribution shift is detected, as in SEMA~\cite{wang2025sema}, so that new tasks do not overwrite the knowledge of previous adapters. However, model size and inference cost grow with the number of adapters, and isolation only protects past knowledge if tasks induce distinct features, which standard CIL benchmarks, with classes assigned to tasks at random, do not guarantee.

We therefore ask whether a single adapter shared by all tasks suffices once its classifier is kept up to date. Rather than proposing a new method, we keep the classifier consistent with the simplest closed-form tools and study the result. We correct the stored statistics through closed-form drift compensation~\cite{he2025semantic}, and keep the feature drift small enough for this correction through feature distillation. Correcting this obsolescence lets one adapter outperform multi-adapter methods with a single forward pass, and a controlled study shows that multiple adapters only help when tasks are clearly distinct.

\section{One adapter with an up-to-date classifier}

\paragraph{Model.}
We keep a pre-trained Vision Transformer (ViT) frozen and add a single adapter shared by all tasks. As in AdaptFormer~\cite{chen2022adaptformer}, the adapter is a small bottleneck MLP, $A(h) = W_{\uparrow}\,\sigma(W_{\downarrow} h)$ with $W_{\downarrow}\in\mathbb{R}^{r\times d}$, $W_{\uparrow}\in\mathbb{R}^{d\times r}$, $r \ll d$ and a ReLU $\sigma$, inserted in parallel with the MLP of each block $l\in\mathcal{B}$. Multi-adapter methods keep adding sets of such modules; we keep one set, whose parameters $\theta_t$ after task $t$ are initialised from $\theta_{t-1}$. An input $x$ is represented by its $\ell_2$-normalised \texttt{[CLS]} embedding $z=\phi_\theta(x)\in\mathbb{R}^d$. The adapter is trained on each task with SGD and a temporary linear head, which is then discarded, and classification is left to an analytic classifier.

\paragraph{Analytic classifier.} Classification is done by ridge regression from the features $z_i$ of the training samples $i$ seen so far to their one-hot labels $y_i$~\cite{fontenla2021rolann}:

\begin{equation}
\label{eq:ridge}
    W = \arg\min_{W} \sum_i \| W^\top z_i - y_i\|_2^2 + \lambda\|W\|_F^2 = (G+\lambda I)^{-1}M,
\end{equation}

\noindent with $G = \sum_i z_i z_i^\top$ and $M = \sum_i z_i y_i^\top$. The solution depends on the data only through these two statistics. Since both are sums over samples, they are updated task by task without storing data, and the incremental solution equals training on all data at once. As classes accumulate, however, the positive samples of each output become a small minority, so in practice we use a class-balanced weighted ridge regression~\cite{zong2013weighted}. Its solution requires the Gram matrix of each class rather than only $G$. To keep the storage small, we approximate each of them by the within-class covariance shared by all classes plus the $q=4$ leading eigenvectors of the class covariance, which are updated by drift compensation like $G$ and $M$.

\paragraph{Classifier obsolescence.} This equivalence, however, holds only if $\phi_\theta$ is fixed. When the adapter keeps learning, $G$ and $M$ contain statistics of features that the model no longer produces, and $W$ becomes \emph{obsolete}. Our goal is therefore to keep $G$ and $M$ expressed in the feature space of the current adapter without revisiting past data. We combine \emph{drift compensation} (Eqs.~\eqref{eq:transport}--\eqref{eq:compensation}), which updates the statistics after each task, with \emph{feature distillation} (Eq.~\eqref{eq:loss}), which keeps this drift small.

When task $t$ ends, $G$ and $M$ describe features produced by $\theta_{t-1}$, whereas the classifier will be applied to features produced by $\theta_t$. Compensation assumes that the drift between both is linear, i.e., $\phi_{\theta_t}(x) = P^\top \phi_{\theta_{t-1}}(x)$ for every input $x$ and some $P\in\mathbb{R}^{d\times d}$. Substituting into $G$ and $M$, the statistics transform exactly as

\begin{equation}
    \label{eq:transport}
    G \leftarrow P^\top G P, \qquad M \leftarrow P^\top M,
\end{equation}

\noindent so they can be updated without access to past data. $P$ is unknown, but it can be estimated from the current data $\mathcal{D}_t$, the only data that can be embedded by both $\theta_{t-1}$ and $\theta_t$. Let $Z_{t-1}, Z_t \in \mathbb{R}^{n \times d}$ be the embeddings of the $n$ samples of $\mathcal{D}_t$ produced by $\theta_{t-1}$ and $\theta_t$, respectively (one sample per row). We fit $P$ by ridge regression towards the identity, a simplified variant of~\cite{he2025semantic}:

\begin{equation}
\label{eq:compensation}
\begin{aligned}
  P &= \arg\min_{P}\ \|Z_{t-1}P - Z_t\|_F^2 + \mu\|P - I\|_F^2 \\
    &= I + \big(Z_{t-1}^\top Z_{t-1} + \mu I\big)^{-1} Z_{t-1}^\top \big(Z_t - Z_{t-1}\big),
\end{aligned}
\end{equation}

\noindent where $\mu = \mathrm{tr}(Z_{t-1}^\top Z_{t-1})/d$ is the average eigenvalue of $Z_{t-1}^\top Z_{t-1}$, so no tuning is required. Because the penalty acts on $P-I$, $P$ departs from the identity only along directions spanned by the current data, so the statistics of past classes are modified only where a drift has been observed. Applying $P$ to past inputs assumes a drift shared by all inputs, plausible since all tasks share one adapter. The cost is one $d \times d$ linear system per task.

Distillation addresses the limits of this compensation, which is exact only for a linear drift and reliable only near the identity; neither holds if the adapter changes freely. We therefore limit the feature drift during training. For $t>1$, the adapter minimises

\begin{equation}
\label{eq:loss}
    \mathcal{L} = \mathcal{L}_{\mathrm{CE}} + \beta\,\mathcal{L}_{\mathrm{D}}, \qquad \mathcal{L}_{\mathrm{D}} = \big\|\phi_{\theta_t}(x) - \phi_{\theta_{t-1}}(x)\big\|_2^2,
\end{equation}

\noindent where $\mathcal{L}_{\mathrm{CE}}$ is the cross-entropy of the temporary head and $\mathcal{L}_{\mathrm{D}}$, averaged over the current batch, penalises the displacement of the embeddings with respect to the frozen $\theta_{t-1}$. For $t=1$, $\mathcal{L} = \mathcal{L}_{\mathrm{CE}}$. Unlike compensation, however, distillation constrains the adapter itself, so $\beta$ trades plasticity for stability.

\section{Experimental Setup}
\label{sec:setup}

All methods use a frozen ViT-B/16 pre-trained on ImageNet-21K. We evaluate on ImageNet-R (200 classes, 10 tasks), CIFAR-100 (100 classes, 20 tasks) and VTAB (50 classes, 5 tasks), without replay and over ten random class orders. We compare with multi-adapter methods, which isolate knowledge in additional adapters (EASE~\cite{zhou2024ease}, SEMA~\cite{wang2025sema}, TUNA~\cite{wang2025integrating}), and with SLDC~\cite{rao2026sldc}, a single-adapter method that, like ours, continually trains one shared adapter. All keep their original training recipes, but insert adapters with bottleneck $r=16$ in blocks $\mathcal{B} = \{4, \dots, 12\}$. As references, we also report the backbone without adapter and an adapter frozen after the first task. Our adapter (dropout 0.1) is trained for 20 epochs per task with SGD (learning rate 0.005), $\beta=1$ and $\lambda=0.1$. We report final accuracy $A_T$, average incremental accuracy $\bar A$, forgetting $F$ (the drop from the best to the final accuracy on the classes of each task, averaged over tasks), and the inference cost as forward passes per test image. Latency (batch size 1) and energy are measured on an NVIDIA RTX 4090. Code is available at \url{https://github.com/AlejandroDopico2/one-adapter-one-pass}.

\section{Results}

\begin{table}[t]
\centering\footnotesize
\setlength{\tabcolsep}{3.3pt}
\begin{tabular}{@{}c@{\hspace{2pt}}lc cc cc cc}
\toprule
&  & & \multicolumn{2}{c}{ImageNet-R (10 tasks)} & \multicolumn{2}{c}{CIFAR-100 (20 tasks)} & \multicolumn{2}{c}{VTAB (5 tasks)} \\
\cmidrule(lr){4-5}\cmidrule(lr){6-7}\cmidrule(lr){8-9}
& Method & Inference & $A_T\uparrow$ & $\bar A\uparrow$ & $A_T\uparrow$ & $\bar A\uparrow$ & $A_T\uparrow$ & $\bar A\uparrow$ \\
\midrule
& No adapter & $1\times$ & 65.3{\scriptsize$\pm$0.0} & 72.7 & 86.2{\scriptsize$\pm$0.0} & 91.0 & 92.8{\scriptsize$\pm$0.0} & 93.5 \\
& Frozen adapter & $1\times$ & 79.3{\scriptsize$\pm$0.3} & 84.0 & 90.2{\scriptsize$\pm$0.5} & 93.9 & 93.7{\scriptsize$\pm$0.3} & 94.2 \\
\midrule
\multirow{3}{*}{\rotatebox{90}{\tiny Multi}} & EASE & $T\times$ & 75.4{\scriptsize$\pm$0.4} & 81.4 & 85.0{\scriptsize$\pm$0.4} & 90.3 & 92.3{\scriptsize$\pm$0.1} & 92.9 \\
& SEMA & $1\times$ & 73.1{\scriptsize$\pm$0.7} & 79.9 & 82.8{\scriptsize$\pm$1.0} & 89.7 & 87.5{\scriptsize$\pm$2.2} & 90.8 \\
& TUNA & $(T{+}1)\times$ & 78.9{\scriptsize$\pm$0.3} & 83.7 & 90.6{\scriptsize$\pm$0.2} & 93.8 & 92.4{\scriptsize$\pm$0.2} & 92.9 \\
\midrule
\multirow{2}{*}{\rotatebox{90}{\tiny Single}} & SLDC & $1\times$ & 79.1{\scriptsize$\pm$0.3} & 83.6 & 90.8{\scriptsize$\pm$0.2} & 93.9 & 93.5{\scriptsize$\pm$0.4} & 93.8 \\
& Ours & $1\times$ & \textbf{80.4{\scriptsize$\pm$0.3}} & \textbf{84.6} & \textbf{91.0{\scriptsize$\pm$0.2}} & \textbf{94.3} & \textbf{94.2{\scriptsize$\pm$0.3}} & \textbf{94.5} \\
\bottomrule
\end{tabular}
\caption{$A_T$ and $\bar A$ (\%, mean{\scriptsize$\pm$}std over ten class orders; $T$: number of tasks). Ours has a significantly higher $A_T$ than every method on every benchmark (Wilcoxon--Holm, $p<0.05$).}
\label{tab:main}
\end{table}

\begin{table}[t]
\centering\scriptsize
\setlength{\tabcolsep}{3.7pt}
\begin{tabular}{lcccccc}
\toprule
Method & Inference & Params (M) & Storage (MB) & GFLOPs & Latency (ms) & Energy (J/1k) \\
\midrule
EASE & $10\times$  & 3.82 & 14.6  & 352.1 & 49.6 & 4754 \\
SEMA & \textbf{1}$\times$   & 2.24 & 8.6   & \textbf{35.2} & 6.1  & 479  \\
TUNA & $11\times$  & 2.89 & 461.6 & 387.3 & 48.8 & 5252 \\
\midrule
SLDC & \textbf{1}$\times$ & \textbf{0.40} & 452.7 & \textbf{35.2} & \textbf{5.3} & 458 \\
Ours & \textbf{1}$\times$ & \textbf{0.40} & \textbf{7.4} & \textbf{35.2} & 5.4 & \textbf{456} \\
\bottomrule
\end{tabular}
\caption{Inference cost on ImageNet-R: parameters and storage beyond the frozen ViT, other columns for the full model; energy per 1,000 images. TUNA and SLDC store one covariance matrix per class.}
\label{tab:cost}
\end{table}

Table~\ref{tab:main} compares all methods. Our single adapter is the most accurate on all three benchmarks. It outperforms EASE and SEMA, so isolation brings no advantage, and TUNA, with one forward pass instead of $T{+}1$ and about $11\times$ less energy (Table~\ref{tab:cost}). It also outperforms SLDC, a single-adapter method that stores one $d\times d$ covariance matrix per class, about $60\times$ our storage.

\begin{figure}[t]
\centering
\includegraphics[width=\textwidth]{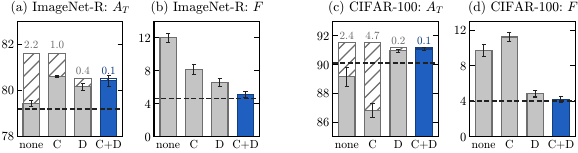}
\caption{Source of forgetting per configuration (none, C: drift compensation, D: feature distillation, C+D: both). (a, c) $A_T$ (solid) and after the oracle refit (hatched). The number above each bar is the obsolescence gap $O$. (b, d) Forgetting $F$. Dashed line: frozen adapter. Mean and standard deviation over three class orders.}
\label{fig:refit}
\end{figure}

Fig.~\ref{fig:refit} analyses whether the forgetting of a continually trained adapter comes from its representation (the features produced by the backbone and the adapter) or from its classifier. To separate both, we compare each configuration with an oracle that refits the classifier on the data of all tasks with the final adapter. This oracle is not possible in CIL, where past data are not stored, but the difference $O = A_{\mathrm{refit}} - A_T$, which we call the \emph{obsolescence gap}, measures the accuracy lost only because the stored statistics are obsolete. Three observations follow. First, adapting without either mechanism improves the representation but not the final accuracy. After refitting, the adapted features improve on the frozen adapter (dashed line) by more than 2 points on ImageNet-R, but an obsolescence gap of 2.2 points leaves the final accuracy barely above it. Second, compensation alone is not enough on long sequences. It halves the gap over the 10 tasks of ImageNet-R, but over the 20 tasks of CIFAR-100 the drift is too large to be captured linearly, and the compensated statistics become worse than uncorrected ones (2.3 points lower). Third, distillation keeps the drift small enough for compensation to be accurate, and compensation then mainly reduces forgetting (by 1.5 points on ImageNet-R). Combined, the obsolescence gap vanishes, and adaptation beats freezing.

\begin{figure}[t]
\centering
\includegraphics[width=0.63\textwidth]{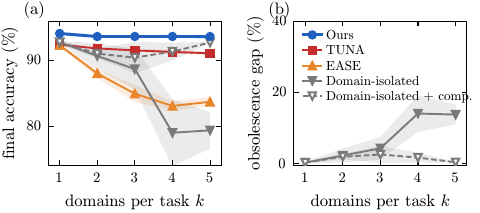}
\caption{VTAB with $k$ domains per task: (a) final accuracy and (b) obsolescence gap of the domain-isolated oracle, with and without compensation (three orders).}
\label{fig:vtab}
\end{figure}

Finally, isolation can only protect past knowledge if each adapter is updated by a single task. We test this on VTAB, which combines five visually distinct domains with ten classes each, so that a natural isolation unit exists. We split the classes of each domain into $k$ groups assigned to different tasks, so that each task mixes $k$ domains while the classes and number of tasks stay the same (Fig.~\ref{fig:vtab}); $k=1$ is the standard setting with one domain per task. As $k$ grows, EASE loses 8.6 points, and a domain-isolated oracle, with one adapter per domain and known domain labels, loses 13.4. Since each domain now appears in several tasks, its adapter is updated several times and the statistics stored for its earlier classes become obsolete. Adding our compensation to the oracle restores its accuracy, so its collapse is caused by classifier obsolescence rather than by a loss of representation quality. Our single adapter barely changes (0.6 points), and neither does TUNA (1.1), whose task adapters are snapshots of a single adapter trained on all tasks. Isolation thus helps only when tasks are clearly separated, which random class orders do not guarantee.

\section{Conclusion}

In class-incremental learning with a pre-trained backbone, much of the forgetting of a continually trained adapter is due to classifier obsolescence rather than to a loss of representation quality. Once this obsolescence is corrected through drift compensation and feature distillation, a single adapter outperforms multi-adapter methods with a single forward pass and a tenth of the energy of TUNA. Isolating knowledge in separate adapters only helps when tasks are clearly distinct; otherwise, even ideal isolation suffers from the same obsolescence. Rather than one adapter per task, new adapters should thus be added only when the single adapter saturates, keeping inference close to one forward pass.

\begin{footnotesize}

\bibliographystyle{abbrv}
\bibliography{bib_short}

\end{footnotesize}

\end{document}